\documentclass[conference]{IEEEtran}
\IEEEoverridecommandlockouts
\usepackage{cite}
\usepackage{amsmath,amssymb,amsfonts}
\usepackage{algorithmic}
\usepackage{graphicx}
\usepackage{textcomp}
\usepackage{booktabs}
\usepackage{xcolor}
\usepackage{url}
\def\BibTeX{{\rm B\kern-.05em{\sc i\kern-.025em b}\kern-.08em
    T\kern-.1667em\lower.7ex\hbox{E}\kern-.125emX}}

\usepackage{siunitx}
\usepackage{tabularx}
\usepackage{booktabs}
\usepackage{float}

\usepackage{url}

\makeatletter 
\newcommand{\linebreakand}{%
  \end{@IEEEauthorhalign}
  \hfill\mbox{}\par
  \mbox{}\hfill\begin{@IEEEauthorhalign}
}
\makeatother 

\begin{document}

\title{Benchmarking Edge Inference Strategies for Deep Learning Models in Industrial Machine Vision
\thanks{The research leading to these results has received funding from “Proyecto Desarrollo de una Estrategia integral de REciCladO de BATerías - RECOBATS”, as part of the TransMisiones 2024 initiative, under project file number PLEC2024-011135. This project is funded by the Spanish Ministry of Science, Innovation and Universities (MICIU), the Spanish State Research Agency (AEI), and the European Regional Development Fund (FEDER, EU), under reference MICIU/AEI/10.13039/501100011033/FEDER, UE, in cooperation with CDTI. This work has also received funding from the Horizon Europe programme under grant agreement No. 101298421 (GRAIL).

Views and opinions expressed are, however, those of the author(s) only and do not necessarily reflect those of the European Union or the European Health and Digital Executive Agency (HADEA). Neither the European Union nor the granting authority can be held responsible for them. }
}

\author{\IEEEauthorblockN{Miguel Gomez Fernandez}
\IEEEauthorblockA{\textit{Smart Systems and Smart Manufacturing (S3M)} \\
\textit{AIMEN}\\
O Porriño, Spain \\
https://orcid.org/0009-0008-0433-9710}

\and
\IEEEauthorblockN{David Castro Boga}
\IEEEauthorblockA{\textit{Smart Systems and Smart Manufacturing (S3M)} \\
\textit{AIMEN}\\
O Porriño, Spain \\
https://orcid.org/0000-0002-1329-770X}

\linebreakand
\IEEEauthorblockN{Roi Mendez-Rial}
\IEEEauthorblockA{\textit{Smart Systems and Smart Manufacturing (S3M)} \\
\textit{AIMEN}\\
O Porriño, Spain \\
https://orcid.org/0000-0002-9991-0316}
\and
\IEEEauthorblockN{Eric Lopez-Lopez}
\IEEEauthorblockA{\textit{Smart Systems and Smart Manufacturing (S3M)} \\
\textit{AIMEN}\\
O Porriño, Spain \\
https://orcid.org/0000-0002-7720-1607}
}

\maketitle

\begin{abstract}
Edge deployment is often the preferred solution for industrial machine vision systems when low latency, data security, or limited connectivity are critical requirements. Several frameworks are available to optimise inference on edge devices; however, relatively few studies have systematically compared their inference-time performance under industrial deployment conditions.

In this work, we present a comparative study of four widely used approaches for machine vision inference in industrial settings: plain PyTorch, ONNX Runtime, OpenVINO, and TensorRT. The evaluation focuses on inference time, covers several CPU- and GPU-based hardware platforms, and includes both conventional convolutional neural networks and a transformer-based vision model. For the hardware platforms and models evaluated, the results show that OpenVINO achieves the lowest inference time on CPUs, while TensorRT achieves the lowest inference time on GPUs. However, TensorRT does not always outperform plain PyTorch for the transformer-based model considered in this study.
\end{abstract}


\section{Introduction}
\label{sec:intro}

Deep learning has strong practical value in industrial machine vision, particularly for tasks such as quality control, precise alignment, and security monitoring \cite{wang2025review, kim2026systematic}. These scenarios often impose strict requirements on latency, data control, and connectivity, making cloud-based inference less appropriate. For this reason, edge deployment is frequently the preferred solution in industrial machine vision systems~\cite{yu2017survey, racano2020performance, chen2019deep, moshsin2025automated}.

However, despite these advantages, edge deployments must operate under tighter constraints in terms of computing power, memory, and energy consumption. As a result, considerable effort is devoted to optimising the inference of a given model for the target hardware. Besides model-level techniques such as quantisation, pruning, or knowledge distillation~\cite{menghani2023efficient, liu2026survey}, inference performance can also be improved through deployment frameworks that provide hardware-specific optimisations and more efficient execution backends. Selecting the most suitable deployment framework is therefore a practical challenge, since its effect on inference time depends on both the hardware platform and the model architecture. 

In contrast to model-level optimisation techniques, the benefits of deployment-framework optimisation are strongly dependent on the target hardware and software stack. As a result, benchmarking studies in this area are often restricted to specific platforms, frameworks, or model families, making it difficult to draw transferable conclusions across deployment scenarios. Among the most widely used deployment frameworks for deep learning inference are PyTorch, ONNX Runtime, OpenVINO, and TensorRT, which provide different execution backends and hardware-specific optimisation capabilities.

Several studies have evaluated inference performance on edge platforms, although most of them focus on a limited subset of frameworks, hardware targets, or neural network architectures. Ulker et al.~\cite{ulker2020reviewing} provided an early broad benchmark across multiple deep learning frameworks and hardware targets. For NVIDIA Jetson platforms, several works have analysed inference performance as a function of the deployment framework~\cite{ratul2025accelerating, aljami2026benchmarking}. Ahn et al.~\cite{ahn2023performance} studied the effects of quantisation across several inference frameworks in a CPU-focused evaluation including both Intel and ARM platforms under MLPerf-style scenarios. In an application-specific edge setting, Surantha et al.~\cite{surantha2025realtime} evaluated multiple frameworks for YOLO-based detection on Raspberry Pi and Jetson devices. However, recent comparative studies jointly covering widely used inference frameworks across heterogeneous CPU- and GPU-based platforms, while also including both CNN-based and transformer-based vision models, remain limited.

To address this gap, this paper benchmarks the inference time of these four widely used deployment frameworks across several hardware platforms relevant to industrial deployment, including both CPU- and GPU-based systems, and considering both CNN-based and transformer-based vision models. The main contributions of this paper are:

\begin{itemize}
    \item A comparative benchmark of PyTorch, ONNX Runtime, OpenVINO, and TensorRT in terms of inference time.
    \item An evaluation across multiple hardware platforms relevant to industrial edge deployment, including both CPU- and GPU-based systems.
    \item A unified evaluation showing how framework performance varies across both CNN-based and transformer-based vision models.
\end{itemize}

The rest of the paper is organised as follows. Section~\ref{sec:methods} describes the benchmark methodology, including the selected hardware platforms, frameworks, and network models. Section~\ref{sec:results} presents and discusses the experimental results. Finally, Section~\ref{sec:conclusions} summarises the main conclusions of the study.

\section{Methodology}
\label{sec:methods}

This section describes the experimental methodology followed in the benchmark study. It includes the hardware platforms used, the inference frameworks considered, the selected deep neural network models, and the procedure employed to measure inference time. The goal is to provide a consistent basis for comparing the behaviour of the evaluated frameworks across different hardware and model configurations.

\subsection{Hardware Platforms}
\label{sec:hardware}

The inference performance study was conducted on the following computing platforms:
\begin{itemize}
    \item \textbf{Machine 1.} A desktop system equipped with an \emph{Intel Core i7-6700} processor operating at $3.40$\,GHz, with $4$ cores and $8$ total threads. The GPU is a \emph{NVIDIA GeForce GTX 1650}.

    \item \textbf{Machine 2.} A desktop system equipped with an \emph{Intel Xeon E5-1650 v4} processor operating at $3.60$\,GHz, with $6$ cores and $12$ total threads. This system includes two GPUs: an \emph{NVIDIA Quadro RTX 6000} and an \emph{NVIDIA GeForce RTX 2080}.

    \item \textbf{Machine 3.} A desktop system equipped with an \emph{Intel Core i9-9820X} processor operating at $3.30$\,GHz, with $10$ cores and $20$ total threads. The GPU is a \emph{NVIDIA GeForce RTX 2080 Ti}.

    \item \textbf{Machine 4.} A \emph{Jetson AGX Orin 32GB} module featuring an $8$-core ARM Cortex-A78AE v8.2 $64$-bit CPU and an integrated NVIDIA Ampere GPU.
\end{itemize}

Since the evaluated platforms differ in several hardware characteristics, the benchmark does not aim to isolate the effect of hardware alone. Instead, the comparison focuses on how each deployment framework behaves within each hardware platform and model configuration.

\subsection{Deployment Frameworks}

The benchmark includes four deployment frameworks: PyTorch, ONNX Runtime, OpenVINO, and TensorRT. These frameworks were selected because they are widely used in practical deep learning deployment and provide different levels of portability and hardware-specific optimisation. All models were exported and executed using FP32 precision. Framework- and hardware-level acceleration modes enabled by default were not manually overridden.

\begin{itemize}
    \item \textbf{PyTorch}~\cite{paszke2019pytorch} serves as the baseline framework, since it is one of the most widely adopted environments for developing and evaluating deep learning models. In this study, it provides the reference implementation against which the other inference frameworks are compared. It supports both CPU and GPU inference.

    \item \textbf{ONNX Runtime}~\cite{onnxruntime} is a general-purpose inference engine for models exported to the ONNX format. Its support for multiple execution providers makes it a flexible option for cross-platform deployment. It supports both CPU- and GPU-based inference.

    \item \textbf{OpenVINO}~\cite{openvino} is an Intel-developed inference framework designed to improve performance on Intel hardware, particularly in CPU-based inference. It supports both CPU and GPU execution, although its GPU backend is limited to Intel GPUs. Since the benchmark does not include Intel GPUs (Section~\ref{sec:hardware}), this study evaluates OpenVINO only on CPUs.

    \item \textbf{TensorRT}~\cite{tensorrt} is a widely used optimisation and inference framework for NVIDIA hardware (NVIDIA GPU's and Jetson AGX Orin). It is specifically designed to improve inference performance through graph optimisations and reduced-precision execution when supported.
\end{itemize}

To reduce the impact of software-version differences on the comparison, the study uses the same framework versions across all devices whenever possible. On the Jetson AGX Orin platform, some versions differ slightly due to platform-specific software constraints. Table~\ref{tab:frameworks} summarises the framework and CUDA versions used on each device.

\begin{table*}
\caption{Framework and CUDA versions used on each device.}
\label{tab:frameworks}
\centering
\begin{tabular}{c c c c c c}
\toprule
Device & CUDA & PyTorch & ONNX Runtime & OpenVINO & TensorRT \\ \midrule
Machine 1 & 12.8 & 2.9.1 & 1.23.2 & 2024.6.0 & 8.6.1 \\
Machine 2 & 12.8 & 2.9.1 & 1.23.2 & 2024.6.0 & 8.6.1 \\
Machine 3 & 12.8 & 2.9.1 & 1.23.2 & 2024.6.0 & 8.6.1 \\
Machine 4 & 11.4 & 2.2.0 & 1.23.2 & 2024.0.0 & 8.5.2 \\
\bottomrule
\end{tabular}
\end{table*}

\subsection{DNNs Used for the Benchmark}

\begin{figure}[t]
    \centering
    \includegraphics[width=0.95\linewidth]{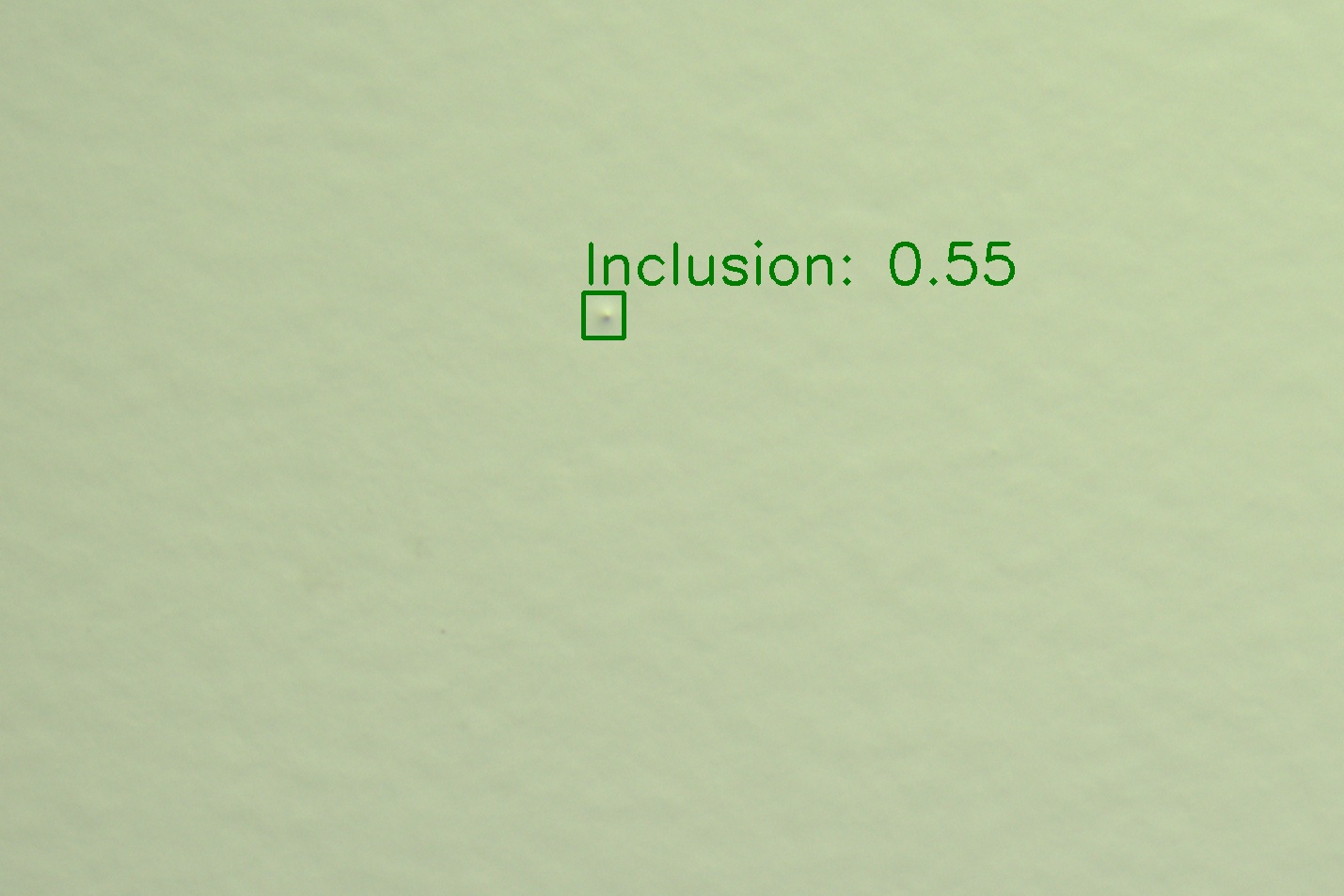}
    \caption{Example of object detection using YOLOv8 for inclusion detection in automatic quality control of painted surfaces \cite{garcia2025coating}.}
    \label{fig:yolo}
\end{figure}

\begin{figure}[t]
    \centering
    \includegraphics[width=0.95\linewidth]{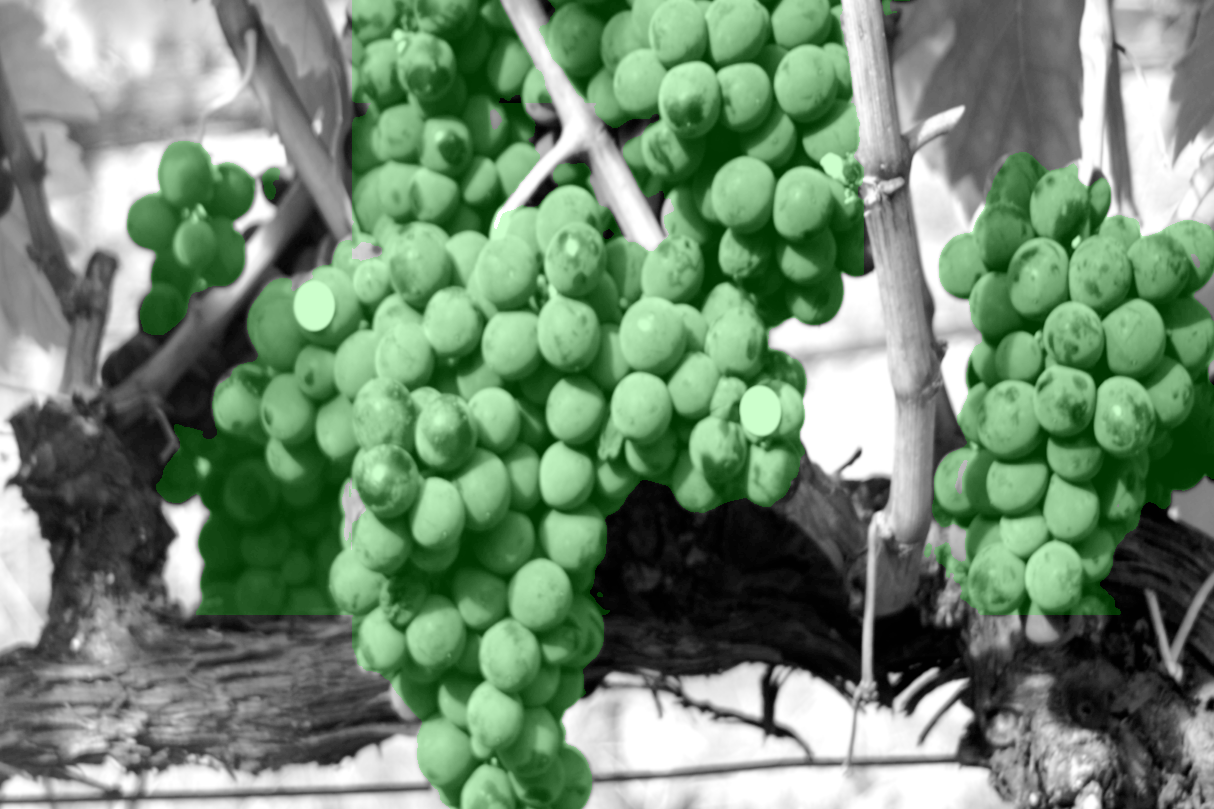}
    \caption{Example of segmentation masks generated by the UNet architecture. The network was trained to segment grapes, highlighted in green, in multispectral images.}
    \label{fig:unet}
\end{figure}

\begin{figure}[t]
    \centering
    \includegraphics[width=0.95\linewidth]{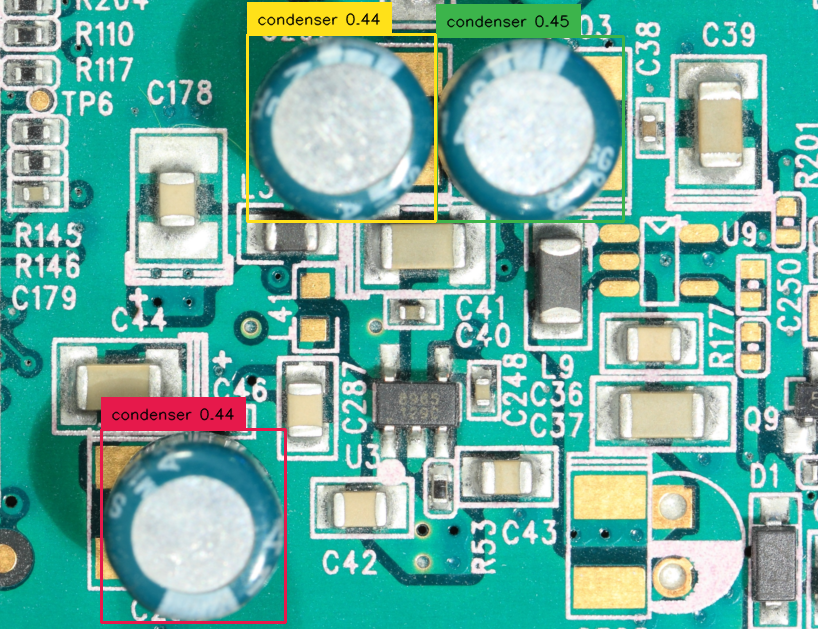}
    \caption{Example of text-guided object detection using Grounding DINO for component detection in PCB manufacturing.}
    \label{fig:gdino}
\end{figure}

The benchmark includes experiments with three deep learning models covering different computer vision tasks and model families, including both CNN-based and transformer-based architectures:

\begin{itemize}
    \item \textbf{YOLOv8}~\cite{varghese2024yolo} is a CNN-based object detection model from the YOLO family, designed for efficient single-stage detection. It was selected as a representative CNN-based architecture for real-time object detection tasks. Fig.~\ref{fig:yolo} shows an example of object detection using YOLOv8.

    \item \textbf{UNet}~\cite{ronneberger2015unet} is a widely used CNN-based architecture for image segmentation, where the output is a pixel-wise prediction map. The backbone is a ResNext50 (32$\times$4d) \cite{xie2017aggregated}. It was selected as a representative model for dense prediction tasks. Fig.~\ref{fig:unet} shows an example of segmentation using UNet.

    \item \textbf{Grounding DINO}~\cite{liu2023grounding} is a transformer-based vision-language object detection model that predicts object locations from textual category names or expressions. It was selected as a representative example of newer text-guided object detection models. Fig.~\ref{fig:gdino} shows an example of object detection using Grounding DINO with a text prompt.
\end{itemize}

\subsection{Experiments}

This benchmark focuses on inference time. It measures inference time for all valid combinations of network, hardware platform, and deployment framework introduced above. The reported measurements include only network inference time and exclude pre-processing and post-processing stages. For each configuration, the benchmark repeats inference 1000 times. The evaluation discards the first 10 runs to avoid initialization and warm-up effects, which significantly increase execution time during the first iterations. It then computes the mean and standard deviation from the remaining 990 runs. All experiments use batch size $1$. Each framework runs the same model variant and input size for a given benchmark case. For GPU-based measurements, the timing procedure includes device synchronization to ensure accurate inference-time estimation.

For the YOLOv8 experiments, the benchmark uses input images of size $640 \times 640 \times 3$. For UNet, it uses input images of size $1024 \times 1024 \times 3$. For Grounding DINO, it uses input images of size $1200 \times 800 \times 3$ with the text prompt \emph{dog}.

\section{Results and Discussion}
\label{sec:results}

\begin{table*}
\caption{YOLOv8 average and standard deviation inference times (ms) across frameworks and machines.}
\label{tab:yolo}
\centering

\begin{tabular}{l r r r r r r r r }
\toprule
& \multicolumn{2}{c}{PyTorch}
& \multicolumn{2}{c}{ONNXRunTime}
& \multicolumn{2}{c}{OpenVINO}
& \multicolumn{2}{c}{TensorRT} \\
& \multicolumn{1}{c}{Mean} & \multicolumn{1}{c}{Std} & \multicolumn{1}{c}{Mean} & \multicolumn{1}{c}{Std} & \multicolumn{1}{c}{Mean} & \multicolumn{1}{c}{Std} & \multicolumn{1}{c}{Mean} & \multicolumn{1}{c}{Std} \\
\midrule
Intel i7-6700 & $78.8\phantom{00}$ & $3.3\phantom{00}$ & $53.7\phantom{00}$ & $2.7\phantom{00}$ & $\mathbf{41.2\phantom{00}}$ & $2.3\phantom{00}$ & {} & {} \\
Intel Xeon E5-1650 v4 & $48.7\phantom{00}$ & $1.7\phantom{00}$ & $38.6\phantom{00}$ & $7.0\phantom{00}$ & $\mathbf{22.65\phantom{0}}$ & $0.81\phantom{0}$ & {} & {} \\
Intel i9-9820X & $28.7\phantom{00}$ & $2.9\phantom{00}$ & $24.1\phantom{00}$ & $1.5\phantom{00}$ & $\mathbf{10.9\phantom{00}}$ & $1.2\phantom{00}$ & {} & {} \\
Jetson AGX Orin (CPU) & $470\phantom{.000}$ & $54\phantom{.000}$ & $\mathbf{184\phantom{.000}}$ & $47\phantom{.000}$ & $429.2\phantom{00}$ & $2.5\phantom{00}$ & {} & {} \\
\midrule
NVIDIA GTX1650 & $8.9\phantom{00}$ & $1.5\phantom{00}$ & $11.42\phantom{0}$ & $0.20\phantom{0}$ & {} & {} & $\mathbf{6.180}$ & $0.037$ \\
NVIDIA Quadro RTX6000 & $7.780$ & $0.063$ & $3.35\phantom{0}$ & $0.13\phantom{0}$ & {} & {} & $\mathbf{1.55\phantom{0}}$ & $0.10\phantom{0}$ \\
NVIDIA RTX2080 & $7.910$ & $0.084$ & $4.340$ & $0.089$ & {} & {} & $\mathbf{2.150}$ & $0.098$ \\
NVIDIA RTX2080Ti & $5.270$ & $0.064$ & $4.750$ & $0.057$ & {} & {} & $\mathbf{2.100}$ & $0.018$ \\
NVIDIA Jetson AGX Orin (GPU) & $20.84\phantom{0}$ & $0.56\phantom{0}$ & $19.69\phantom{0}$ & $0.16\phantom{0}$ & {} & {} & $\mathbf{10.57\phantom{0}}$ & $0.15\phantom{0}$ \\

\bottomrule
\end{tabular}
\end{table*}

\begin{figure}[t]
    \centering
    \includegraphics[width=\linewidth]{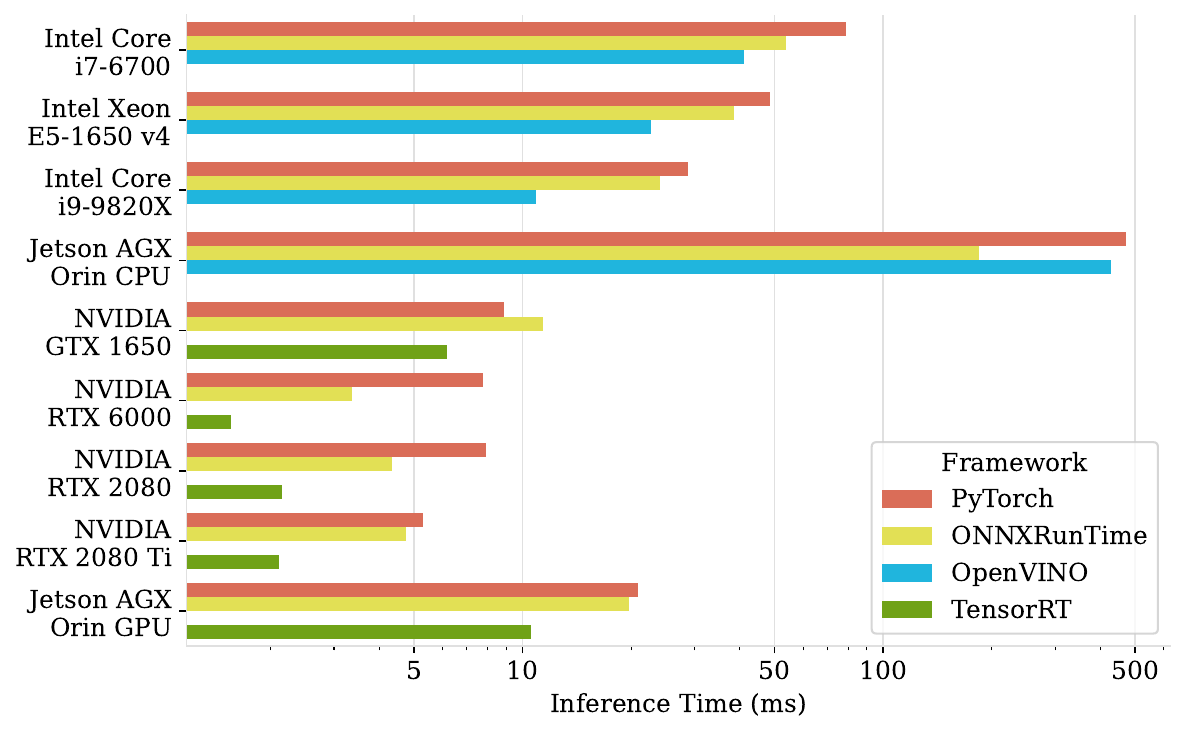}
    \caption{YOLOv8 inference execution time (ms) shown on a logarithmic scale.}
    \label{fig:yolo_gr}
\end{figure}

\begin{figure}[t]
    \centering
    \includegraphics[width=\linewidth]{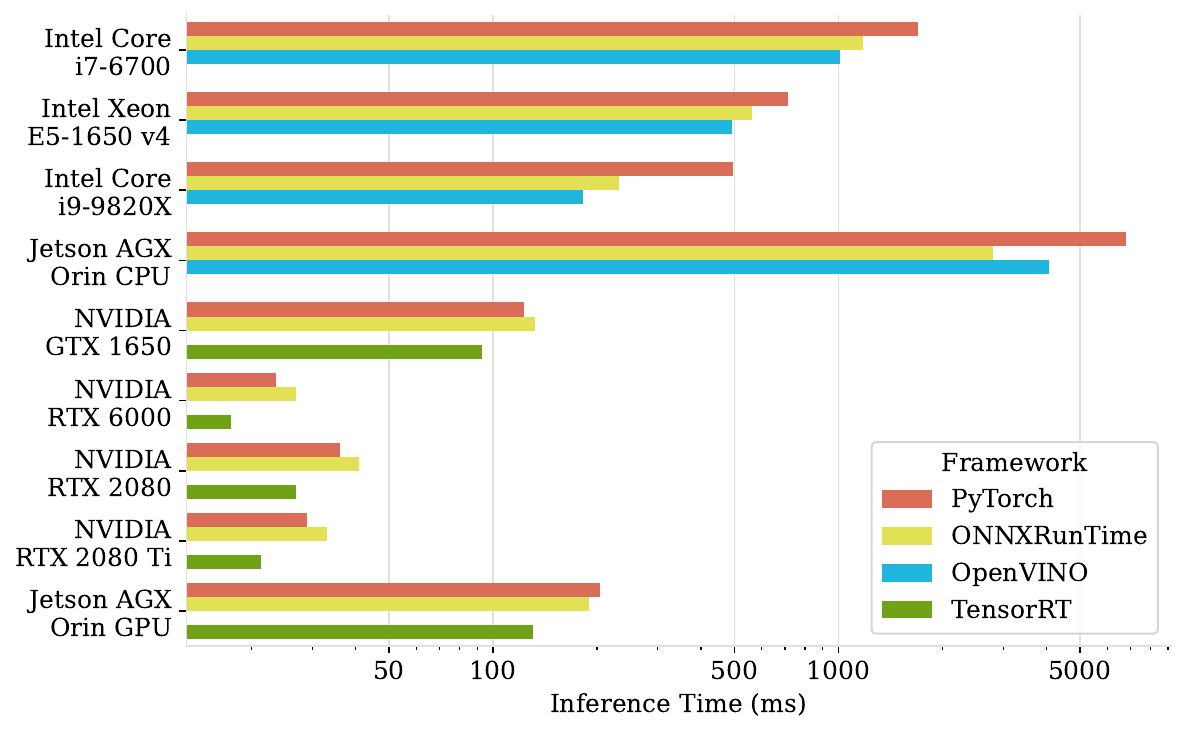}
    \caption{UNet inference execution time (ms) shown on a logarithmic scale.}
    \label{fig:unet_gr}
\end{figure}

\begin{figure}[t]
    \centering
    \includegraphics[width=\linewidth]{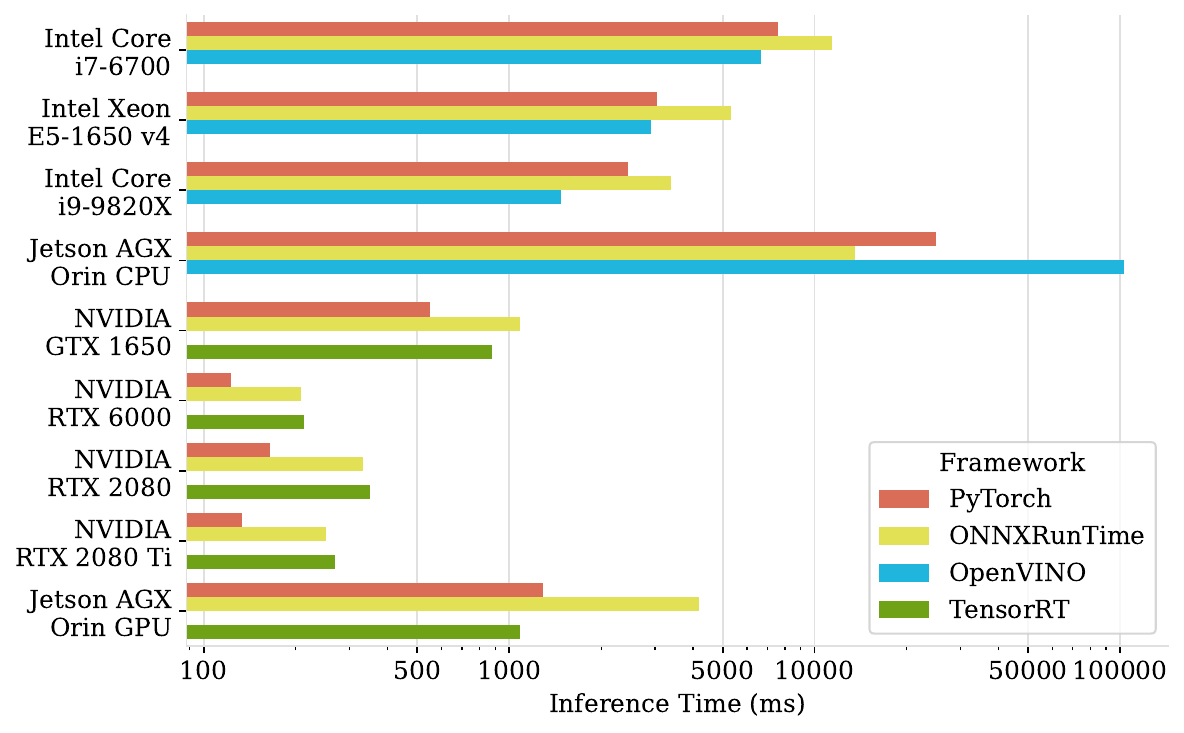}
    \caption{Grounding DINO inference execution time (ms) shown on a logarithmic scale.}
    \label{fig:gdino_gr}
\end{figure}

As discussed in the methodology, this benchmark primarily aims to compare inference frameworks within the same hardware and model configuration. Since the evaluated platforms are heterogeneous, the study does not fully isolate the effect of hardware alone, and direct comparisons across machines must therefore be interpreted with caution. Accordingly, the results are analysed first from the perspective of framework differences across hardware configurations, and then from the perspective of framework behaviour across the tested models. The measured inference times are reported in Tables~\ref{tab:yolo}--\ref{tab:gdino} (bold values highlight the minimum inference time obtained for a given hardware configuration) and Figures~\ref{fig:yolo_gr}--\ref{fig:gdino_gr}.

For CPU inference, the three Intel-based desktop platforms show a consistent pattern across the evaluated models. OpenVINO provides the lowest inference time in all three cases, ONNX Runtime usually ranks second, and PyTorch is the slowest option. This advantage is especially clear for YOLOv8, where OpenVINO reduces inference time by $30$--$60$\% on Machines 1--3. The same trend appears for the other models, although the absolute latencies are higher than with YOLO, especially for the Grounding DINO case. 

The Jetson CPU shows a different behaviour. For the three tested models, ONNX Runtime provides the lowest inference time, while OpenVINO loses the advantage observed on the Intel-based platforms. The difference is moderate for YOLOv8 and UNet, but becomes more notorious for Grounding DINO, where OpenVINO reaches $102720$\,ms, compared with $13580$\,ms for ONNX Runtime, which is more than seven times higher. This is the largest framework-dependent gap observed in the benchmark and highlights the strong dependence of framework performance on the target hardware.

For GPU inference, YOLOv8 and UNet show a clear and consistent pattern across the evaluated NVIDIA platforms. TensorRT provides the lowest inference time in all tested cases, both on the desktop GPUs and on the Jetson GPU. For YOLOv8, TensorRT reduces latency by roughly $30$--$56$\% with respect to the next best framework. For UNet, the improvement is smaller but still consistent, with TensorRT reducing latency by about $25$--$31$\%. These results indicate that TensorRT provides the most effective optimisation path for GPU deployment of the CNN-based models considered in this study.

Grounding DINO exhibits a different behaviour on GPUs (Table~\ref{tab:gdino}). On the desktop GPUs, PyTorch provides the lowest inference time in all tested cases, while TensorRT does not improve over the baseline and is between approximately $1.6$ and $2.1$ times slower. Moreover, TensorRT only slightly outperforms ONNX Runtime on the GTX1650. The Jetson GPU partially breaks this pattern, since TensorRT performs substantially better than PyTorch ($1082.7$\,ms versus $1290.8$\,ms), although the absolute latency remains much higher than on the desktop GPUs. Overall, these results indicate that the GPU-side advantages observed with TensorRT for CNN-based models do not necessarily transfer to transformer-based vision-language models under the same export and deployment pipeline. This behaviour is likely influenced by the structure of the exported Grounding DINO graph, its dynamic text inputs, and the hardware-specific optimizations selected by each runtime.

A general comparison across models reinforces this point. For YOLOv8 and UNet (Tables~\ref{tab:yolo} and \ref{tab:unet}, respectively), the ranking of frameworks is stable: OpenVINO performs best on the Intel-based CPUs, ONNX Runtime performs best on the Jetson CPU, and TensorRT performs best on GPUs. Grounding DINO changes this picture mainly on GPU inference, where PyTorch becomes the strongest desktop option and TensorRT loses the advantage observed for the CNN-based models. Therefore, the benchmark suggests that framework selection depends not only on the target hardware, but also on the model family. In the evaluated setting, hardware-specific runtimes provide clear benefits for CNN-based inference, whereas transformer-based vision-language models show a less favourable response to those same optimisation paths.

\begin{table*}
\caption{UNet average and standard deviation inference times (ms) across frameworks and machines.}
\label{tab:unet}
\centering

\begin{tabular}{l r r r r r r r r}
\toprule
& \multicolumn{2}{c}{PyTorch}
& \multicolumn{2}{c}{ONNXRunTime}
& \multicolumn{2}{c}{OpenVINO}
& \multicolumn{2}{c}{TensorRT} \\
& \multicolumn{1}{c}{Mean} & \multicolumn{1}{c}{Std} & \multicolumn{1}{c}{Mean} & \multicolumn{1}{c}{Std} & \multicolumn{1}{c}{Mean} & \multicolumn{1}{c}{Std} & \multicolumn{1}{c}{Mean} & \multicolumn{1}{c}{Std} \\
\midrule
Intel i7-6700 & $1696\phantom{.000}$ & $22\phantom{.000}$ & $1175\phantom{.000}$ & $25\phantom{.000}$ & $\mathbf{1008\phantom{.000}}$ & $59\phantom{.000}$ & {} & {} \\
Intel Xeon E5-1650 v4 & $714.1\phantom{00}$ & $7.4\phantom{00}$ & $562\phantom{.000}$ & $80\phantom{.000}$ & $\mathbf{492.1\phantom{00}}$ & $3.1\phantom{00}$ & {} & {} \\
Intel i9-9820X & $497\phantom{.000}$ & $19\phantom{.000}$ & $232\phantom{.000}$ & $13\phantom{.000}$ & $\mathbf{182\phantom{.000}}$ & $13\phantom{.000}$ & {} & {} \\
Jetson AGX Orin (CPU) & $6786\phantom{.000}$ & $253\phantom{.000}$ & $\mathbf{2800\phantom{.000}}$ & $99\phantom{.000}$ & $4076\phantom{.000}$ & $19\phantom{.000}$ & {} & {} \\
\midrule
NVIDIA GTX1650 & $123.10\phantom{0}$ & $0.16\phantom{0}$ & $132.72\phantom{0}$ & $0.45\phantom{0}$ & {} & {} & $\mathbf{92.92\phantom{0}}$ & $0.16\phantom{0}$ \\
NVIDIA Quadro RTX6000 & $23.48\phantom{0}$ & $0.19\phantom{0}$ & $26.30\phantom{0}$ & $0.28\phantom{0}$ & {} & {} & $\mathbf{17.47\phantom{0}}$ & $0.22\phantom{0}$ \\
NVIDIA RTX2080 & $36.1\phantom{00}$ & $1.4\phantom{00}$ & $41.0\phantom{00}$ & $1.2\phantom{00}$ & {} & {} & $\mathbf{26.8\phantom{00}}$ & $1.4\phantom{00}$ \\
NVIDIA RTX2080Ti & $29.020$ & $0.071$ & $33.03\phantom{0}$ & $0.16\phantom{0}$ & {} & {} & $\mathbf{21.35\phantom{0}}$ & $0.10\phantom{0}$ \\
NVIDIA Jetson AGX Orin (GPU) & $203.9\phantom{00}$ & $8.6\phantom{00}$ & $190.21\phantom{0}$ & $0.67\phantom{0}$ & {} & {} & $\mathbf{130.77\phantom{0}}$ & $0.21\phantom{0}$ \\

\bottomrule
\end{tabular}
\end{table*}

\begin{table*}
\caption{Grounding DINO average and standard deviation inference times (ms) across frameworks and machines.}
\label{tab:gdino}
\centering

\begin{tabular}{l r r  r r  r r r r}
\toprule
& \multicolumn{2}{c}{PyTorch}
& \multicolumn{2}{c}{ONNXRunTime}
& \multicolumn{2}{c}{OpenVINO}
& \multicolumn{2}{c}{TensorRT} \\
& \multicolumn{1}{c}{Mean} & \multicolumn{1}{c}{Std} & \multicolumn{1}{c}{Mean} & \multicolumn{1}{c}{Std} & \multicolumn{1}{c}{Mean} & \multicolumn{1}{c}{Std} & \multicolumn{1}{c}{Mean} & \multicolumn{1}{c}{Std} \\
\midrule
Intel i7-6700 & $7560\phantom{.00}$ & $190\phantom{.00}$ & $11373\phantom{.00}$ & $39\phantom{.00}$ & $\mathbf{6660\phantom{.00}}$ & $100\phantom{.00}$ & {} & {} \\
Xeon E5-1650 v4 & $3044\phantom{.00}$ & $19\phantom{.00}$ & $5330\phantom{.00}$ & $470\phantom{.00}$ & $\mathbf{2920.3\phantom{0}}$ & $7.6\phantom{0}$ & {} & {} \\
Intel i9-9820X & $2450\phantom{.00}$ & $170\phantom{.00}$ & $3388\phantom{.00}$ & $55\phantom{.00}$ & $\mathbf{1479\phantom{.00}}$ & $27\phantom{.00}$ & {} & {} \\
Jetson AGX Orin (CPU) & $24964\phantom{.00}$ & $305\phantom{.00}$ & $\mathbf{13580\phantom{.00}}$ & $310\phantom{.00}$ & $102720\phantom{.00}$ & $290\phantom{.00}$ & {} & {} \\
\midrule
NVIDIA GTX1650 & $\mathbf{549.67}$ & $0.74$ & $1082.34$ & $0.60$ & {} & {} & $878.48$ & $0.34$ \\
NVIDIA Quadro RTX6000 & $\mathbf{123.0\phantom{0}}$ & $1.5\phantom{0}$ & $207.4\phantom{0}$ & $2.2\phantom{0}$ & {} & {} & $213.2\phantom{0}$ & $2.8\phantom{0}$ \\
NVIDIA RTX2080 & $\mathbf{165.2\phantom{0}}$ & $7.4\phantom{0}$ & $331\phantom{.00}$ & $14\phantom{.00}$ & {} & {} & $350\phantom{.00}$ & $19\phantom{.00}$ \\
NVIDIA RTX2080Ti & $\mathbf{133.5\phantom{0}}$ & $1.3\phantom{0}$ & $250.4\phantom{0}$ & $2.9\phantom{0}$ & {} & {} & $268\phantom{.00}$ & $10\phantom{.00}$ \\
NVIDIA Jetson AGX Orin (GPU) & $1290.8\phantom{0}$ & $3.7\phantom{0}$ & $4170\phantom{.00}$ & $200\phantom{.00}$ & {} & {} & $\mathbf{1082.7\phantom{0}}$ & $1.2\phantom{0}$ \\
\bottomrule
\end{tabular}
\end{table*}

\section{Conclusions}
\label{sec:conclusions}

This work presented an inference-time benchmark of four deployment frameworks (PyTorch, ONNX Runtime, OpenVINO and TensorRT) across several hardware platforms and three representative deep learning models: an object detector, a segmentation model, and a transformer-based vision-language model.

The results show a consistent pattern for CPU inference on the Intel-based desktop platforms, where OpenVINO provides the lowest inference times for all three evaluated models. In contrast, on the Jetson CPU, ONNX Runtime yields the best results, which is consistent with the fact that OpenVINO is primarily optimised for Intel hardware. For GPU inference, the observed behaviour depends more strongly on the model family. TensorRT provides the best results for the CNN-based models, namely YOLOv8 and UNet, across the evaluated NVIDIA GPUs. However, this advantage does not cleanly extend to the transformer-based model considered in this study, Grounding DINO, for which PyTorch tends to achieve the lowest inference times on the desktop GPUs used in this benchmark.

Overall, the benchmark shows that framework selection depends not only on the target hardware, but also on the model architecture. In the evaluated setting, hardware-specific runtimes provide clear benefits for CNN-based inference, while their advantages are less consistent for transformer-based vision-language models.

\bibliographystyle{IEEEtran}
\bibliography{IEEEabrv,references}

\end{document}